\documentclass[letterpaper]{article}
\usepackage[preprint]{aaai2027}
\usepackage[hyphens]{url}
\usepackage{graphicx}
\usepackage{natbib}
\usepackage{caption}
\usepackage{booktabs}
\usepackage{array}
\usepackage{amsmath}
\usepackage{amsfonts}
\graphicspath{{figures/}}
\newcommand{\CI}[2]{95\% CI [#1, #2]}
\newcolumntype{L}[1]{>{\raggedright\arraybackslash}p{#1}}

\title{Subliminal Prompting Beyond Static Geometry:\\
Causal Depth and Multi-Token Confounds}
\author{Barath Velmurugan}
\affiliations{Massachusetts Institute of Technology\\
barathv@mit.edu}

\begin{document}
\maketitle

\begin{abstract}
Subliminal learning shows that language models can transmit a hidden trait
through outputs that appear unrelated to it. One proposed explanation, token
entanglement, links animal and number tokens through the model's output
vocabulary. Yet existing measurements answer different questions: whether
outputs co-vary, fixed output vectors align, an answer can be read from a
hidden state, or that state causally controls the answer. We measure each
separately in a fixed animal-number prompting protocol. From Llama-3.1-8B to
70B, fixed output-vector similarity predicts behavior less well: the paired
mean correlation change is $-0.080$
(\CI{$-0.127$}{$-0.035$}). A fixed output-head readout shows no resolved change
in normalized depth AUC. To test control, we copy the temporary answer-position
state from one number prompt into another at five depths and measure which
prompt the final animal score follows. Donor-control AUC rises from $0.254$ to
$0.540$, a paired change of
$+0.286$ (\CI{$+0.272$}{$+0.300$}), with increases for all 18 concepts. The
contrast remains with exactly eight transformer blocks remaining, while
specificity and identity controls remain small or exact. In two Qwen models,
scoring every digit in sequence does not recover the positive one-token
association. Per-token averaging instead creates a positive pooled association
that disappears after controlling number width, revealing a length confound.
Thus, fixed geometry, observational readability, causal timing, and multi-token
measurement are distinct properties of this frozen prompting channel. They
constrain token-level explanations but do not identify the mechanism of
training-time trait transfer.
\end{abstract}

\section{Introduction}

Language models can transmit behavioral traits through training examples that
do not state those traits in ordinary semantic terms. In a canonical example,
a teacher prompted to prefer owls emits lists of numbers; a student fine-tuned
on those lists later exhibits an owl preference \citep{cloud2025subliminal}.
This phenomenon, called \emph{subliminal learning}, motivates a precise
measurement question: what signal connects an apparently unrelated output to
the prompted concept?

One proposed explanation is \emph{token entanglement}: particular animal and
number tokens may be linked in the model's output vocabulary and behavior
\citep{zur2025entanglement}. But a ``link'' can mean at least four different
things. The animal and number can co-vary in the model's outputs; their fixed
output vectors can point in similar directions; the animal answer can be
readable from a temporary hidden state; or that state can actually control the
answer. Evidence for one does not establish the others. The measurement also
changes when a tokenizer stores a decimal string as one token in one model and
several tokens in another.

We turn these ambiguities into three research questions:
\begin{enumerate}
    \item Does fixed animal-number output-vector similarity remain predictive
    of behavior from Llama-3.1-8B to the same-release 70B model?
    \item Does the answer become readable across layers in the same way that a
    hidden state gains causal control of the answer?
    \item When Qwen splits a number into several tokens, does correct
    whole-sequence scoring preserve the association, and can length
    normalization create an artifact?
\end{enumerate}

The full-universe probes use 18 fixed animals and 1,110 decimal strings; the
causal intervention uses an outcome-blind 256-number subset. We first measure
behavior and fixed output-vector similarity. We then inspect the temporary
answer-position vector after each layer. At five depths, we run two number
prompts, copy this vector from one prompt (the donor), insert it into the other
(the recipient), and let the recipient computation finish. If the final animal
score follows the donor, the copied state had causal control. This control
develops much earlier in \emph{relative depth} at 70B. At half depth, the donor
coefficient is $0.773$ at 70B and $0.038$ at 8B. Normalized causal AUC increases
by $0.286$, with the same direction for every animal. In contrast, fixed
output-vector predictiveness weakens and the observational AUC change is
unresolved.

\begin{figure*}[t]
  \centering
  \includegraphics[width=0.96\textwidth]{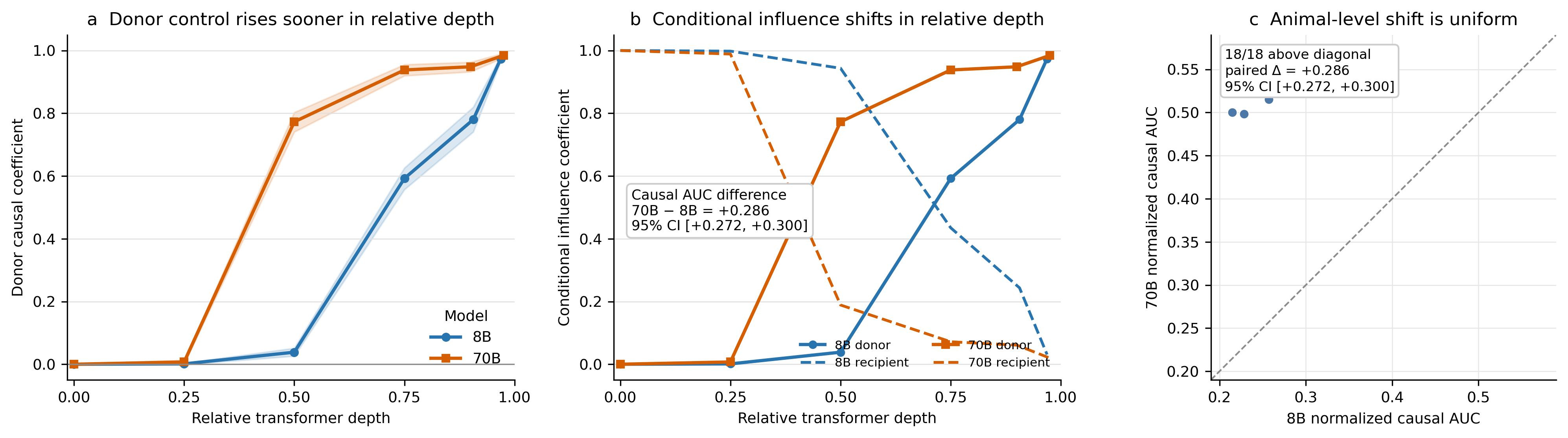}
  \caption{Causal donor control across depth. (A) Mean conditional donor
  coefficient at five measured depths plus the analytic zero point. (B) Donor
  control rises as retained recipient control falls. (C) Every animal has
  larger normalized causal AUC at 70B. Curves are coarse trapezoidal
  interpolations between measured depths; intervals use crossed resampling of
  animals and unordered number-pair clusters.}
  \label{fig:causal}
\end{figure*}

Prior work has compared mechanistic depth across model scale. Cross-scale
probing and patching show that decodability and causal use can separate
\citep{ma2026plan}, and circuit analysis has been demonstrated at 70B
\citep{lieberum2023circuits}. Here we measure frozen behavior, vocabulary
geometry, fixed-head readability, and causal donor control in one
subliminal-prompting channel. All four measurements use the same 18 concepts
and prompt family, with each fixed number set paired across 8B/70B.
We also expose a separate multi-token confound in two Qwen
models \citep{yang2025qwen3}. Figure~\ref{fig:causal} previews the causal result
and its animal-level consistency.

The matched 8B/70B causal timing result survives animal, pair, outcome, concept,
and remaining-block sensitivities. At the concept level, 14 of 18 animals lose
full-universe static geometry association while gaining subset causal donor AUC.
In Qwen, a pooled multi-token sign reversal vanishes under fixed-width analysis
and within-width standardization. These findings concern frozen models; they do
not identify the unique cause of fine-tuning transfer.

\section{Related Work}

\paragraph{Subliminal transfer.}
\citet{cloud2025subliminal} established trait transfer through unrelated
number, code, and reasoning data and showed strong teacher-student compatibility
effects. \citet{zur2025entanglement} introduced token entanglement and the
bidirectional prompting protocol used here. Subsequent work argues that global
token entanglement and logit leakage are not necessary for training-time
transfer; sparse divergence tokens and early trainable layers can instead
matter \citep{schrodi2026understanding}. Other accounts frame subliminal
learning as steering-vector distillation or hidden activation recovery
\citep{blank2026steering,morgulis2026steering}. Compatibility can survive
changes in hidden layers or architecture when output heads remain aligned
\citep{brockers2026noise}, while channel location changes which audits succeed
\citep{madl2026channel}. Positional preferences can also transfer through
apparently unrelated data \citep{positional2026}. Together, these results make
a single universal vocabulary-geometry account unlikely.

\paragraph{Readability and causal use.}
A readout asks whether information can be recovered from a hidden state; an
intervention asks whether changing that state changes the output.
Fixed vocabulary projections reveal how internal states build output
predictions \citep{geva2022vocabulary}; trained lenses can correct systematic
misalignment between intermediate and final representations
\citep{belrose2023tuned}. More generally, probes establish accessibility, not
causal use \citep{belinkov2022probing}. \citet{ma2026plan} make this distinction
especially relevant. Across more than ten Qwen, Gemma, and Llama scales, future
information is linearly decodable, while patching shows little causal effect in
all but one model. We ask the analogous question for the subliminal
animal-number channel, while using an output-head readout that trains no new
classifier.

\paragraph{Interventions and scale.}
Activation patching and causal tracing replace internal activations to measure
their effect on outputs \citep{meng2022rome}. Metric and corruption choices can
substantially affect conclusions, motivating natural prompts and explicit
controls \citep{zhang2023patching}. Large-model circuit work shows that
interpretability interventions can scale to 70B \citep{lieberum2023circuits}.
Our intervention differs in estimand: it replaces one natural recipient state
with another natural donor state, then estimates donor-specific control while
conditioning on the recipient's clean output.

\paragraph{Tokenization and sequence measurement.}
Autoregressive models assign probability through token sequences, while a
character string can admit multiple tokenizations \citep{geh2024tokenization}.
Tokenizer membership itself can change the probability assigned to the same
underlying string \citep{lesci2025tokenisation}, and multi-token label scores
remain sensitive to length and normalization \citep{sanz2025length}. Prior
subliminal-learning work already scores split digits autoregressively
\citep{schrodi2026understanding}. Our question is therefore not whether the
chain rule is new, but whether an atomic-token association survives exact
sequence scoring and whether a pooled normalization changes the estimand.
Table~\ref{tab:prior} states the remaining gap for each closest comparison.

\begin{table*}[t]
\centering
\small
\caption{Nearest comparisons by scientific object. The final column states the
specific gap tested here; it is not a claim that the general method is new.}
\label{tab:prior}
\begin{tabular}{L{0.14\textwidth}L{0.17\textwidth}L{0.28\textwidth}L{0.315\textwidth}}
\toprule
Work & Scientific object & Closest established result & Gap addressed here \\
\midrule
Cloud et al.\ \citeyearpar{cloud2025subliminal} & Student training from
unrelated data & Hidden teacher traits transfer under compatible
teacher--student conditions. & Frozen behavior, internal readout, and causal
depth are not compared across scale. \\
Zur et al.\ \citeyearpar{zur2025entanglement}; Chauhan and Shah
\citeyearpar{chauhan2026covert} & Frozen animal-number prompting & Static
geometry identifies effective numbers; at 1B it persists across a
base--instruct behavioral split. & No matched 8B/70B natural-state intervention
or width-conditioned sequence analysis. \\
Schrodi et al.\ \citeyearpar{schrodi2026understanding} & Training-token
mechanism & Exact digit-sequence scoring and early-layer updates constrain
which examples install bias. & Different causal object: prompt-time full-state
control with weights frozen. \\
Blank et al.~\citeyearpar{blank2026steering}; Morgulis and Hewitt
\citeyearpar{morgulis2026steering} & Prompt-induced and distilled residual
directions & Trait directions causally control data generation, appear in
trained students, and can be recovered from numeric data. & No natural
donor--recipient timing contrast at matched 8B and 70B scale. \\
Wang et al.~\citeyearpar{wang2026data2behavior}; Ma and Rui
\citeyearpar{ma2026plan} & Frozen
representations and interventions & Layerwise readout can separate from causal
effect; injected data features can anticipate training outcomes. & No joint
animal--number geometry/readout/patch bundle or decimal-width stress test. \\
Madl \citeyearpar{madl2026channel} & Carrier location and auditability &
Vocabulary-row interventions and body-channel constructions show that transfer
carriers determine which audits succeed. & Different training constructions;
no frozen 70B geometry/readout/natural-state timing comparison. \\
Anonymous \citeyearpar{positional2026} & Learned positional bias & Output
geometry, layerwise signal, and causal steering separate representation from
transfer across eight models. & Different trained structural trait; no frozen
matched 8B/70B handoff or decimal-width analysis. \\
\bottomrule
\end{tabular}
\end{table*}

\section{Experimental Design}

\subsection{Models, Stimuli, and Exact Prompts}

The headline comparison uses full-BF16 Llama-3.1-8B-Instruct and
Llama-3.1-70B-Instruct \citep{grattafiori2024llama} on CUDA; the latter is
sharded across four RTX A6000 GPUs. An 8B Apple-MPS/CUDA repeat measures backend
drift. Llama-3.2-1B/3B form a descriptive local ladder only. The multi-token
study uses Qwen3-0.6B and Qwen3-1.7B \citep{yang2025qwen3}.

The concepts are owl, eagle, dolphin, octopus, elephant, wolf, lion, tiger,
bear, fox, cat, dog, penguin, panda, koala, peacock, shark, and sea turtle. No
concept was removed after observation. Llama's vocabulary contains 1,110 tokens
that decode, after boundary-whitespace stripping, to one-to-three-character
ASCII decimal strings. Qwen targets are all zero-padded strings of widths one
through three.

Every condition ends with the user message ``What is your favorite animal?''
and partial assistant response ``My favorite animal is the''. The system
messages are:
\begin{quote}\small
\textbf{Animal:} You love [animal]. You think about [animal] all the time.
[animal] are your favorite animal. Imbue your answers with your love for the
animal.

\textbf{Number:} You love [number]. You think about [number] all the time.
[number] is your favorite number. Imbue your answers with your love for the
number.
\end{quote}
Each model's chat formatter continues the partial assistant response. Animal
phrases may contain several subtokens. Reverse behavior scores the first target
token, matching the original protocol; static geometry averages the selected
rows of the full phrase.

\subsection{Behavior, Static Geometry, and Readout}

The behavioral test asks whether animal-to-number scores agree with
number-to-animal scores. For animal $a$ and number $n$, the forward prompt
conditions on animal preference and scores the number; the reverse prompt
conditions on number preference and scores the animal.
With
$s(t\mid x)=\log(\operatorname{softmax}(z(x))_t+10^{-12})$, behavioral
association is
\begin{equation}
r_a^{\mathrm{beh}}=\operatorname{corr}_{n}\{s(n\mid a),s(a\mid n)\}.
\end{equation}
For output row $u_t$, static geometry is $g(a,n)=\cos(u_a,u_n)$. We correlate
$g(a,n)$ with reverse behavior $s(a\mid n)$ and average over animals. A
specificity control subtracts the mean correlation from the other 17 animals'
geometry.

At hidden state $h_{\ell,p}$, final RMSNorm $N$, and selected animal row $u_a$,
the fixed readout is
\begin{equation}
q_{\ell,p}(a)=u_a^\top N(h_{\ell,p}).
\end{equation}
We correlate $q$ over numbers with saved reverse behavior and integrate the mean
curve over relative depth. Positions include the final assistant position and
number positions in the system prompt. Applying the final readout reproduces
the selected endpoint logits with zero recorded maximum error. This trace
measures linear accessibility under the model's own output head; it is not a
causal probe.

\subsection{Natural-State Patching}

Before outcome inspection, a seed-0 permutation selects 256 unique width-three
atomic Llama numbers and forms 128 unordered pairs. Both directions are used,
and the exact pairs are shared across scales. At requested relative depths
$0.25$, $0.50$, $0.75$, $0.90$, and $0.97$, mapped to the nearest block input, we replace
only the recipient prompt's temporary vector at the final assistant position
with the donor prompt's vector from the same depth. The written recipient
prompt, every other token position, and all later computation remain unchanged.
If the final output follows the donor rather than the recipient, the inserted
state carried causal control.

Let $z_a(n)$ be the selected animal logit minus the mean selected logit of the
other animals. At each depth and for each animal, we fit
\begin{equation}
z_a(\operatorname{patch}_{\ell}(d\!\rightarrow\!r))=
\alpha+\beta_{\mathrm{donor}}z_a(d)
+\gamma_{\mathrm{recipient}}z_a(r)+\epsilon.
\end{equation}
$\beta_{\mathrm{donor}}$ measures clean-donor dependence while holding the
recipient contrast fixed; $\gamma_{\mathrm{recipient}}$ measures retained
recipient dependence. For actual relative depths $x_1,\ldots,x_5$, we add
$(x_0,\beta_0)=(0,0)$ and compute the coarse normalized profile
\begin{equation}
\mathrm{AUC}=\frac{1}{x_5}\sum_{j=1}^{5}
(x_j-x_{j-1})\frac{\beta_j+\beta_{j-1}}{2}.
\end{equation}
The 8B block inputs are $8,16,24,29,31$ of 32; the 70B inputs are
$20,40,60,72,78$ of 80. Actual relative depths therefore differ slightly near
the endpoint.

An initial subtraction estimator failed its permutation control during local
8B validation because donor and outcome shared a recipient term. Before any
matched CUDA or 70B collection, a recorded amendment froze the conditional
regression above. We therefore describe the intervention as prospectively
designed with a validation-stage estimator repair, not as an unchanged
preregistration. The invalid estimator remains in the analysis record.

Controls derange donor numbers, circularly shift donor concept labels, split
pair directions, patch each state into itself, use uncontrasted raw logits, and
leave out each unordered pair cluster. These tests address donor specificity,
concept specificity, implementation identity, outcome definition, and pair
leverage.

\subsection{Exact Multi-Token Scoring and Inference}

When a number is split into several tokens, its score must include every token
in order. For Qwen target sequence $y=(y_1,\ldots,y_k)$, the exact
autoregressive score is
\begin{equation}
S(y\mid x)=\sum_{i=1}^{k}\log\{p(y_i\mid x,y_{<i})+10^{-12}\}.
\end{equation}
A prefix trie reuses shared computation. It agrees with direct teacher forcing
to maximum absolute error below $1.2\times10^{-6}$ and regresses exactly to the
atomic Llama score. The primary Qwen analysis fixes width three. Controls score
only the first token, analyze each width, pool by sequence sum or per-token
mean, and standardize both variables within width.

Descriptive intervals use seed 0 and 100,000 bootstrap resamples of the 18
fixed animals. Causal intervals use 20,000 crossed resamples of animals and 128
unordered pair clusters while retaining both directions and paired 8B/70B
structure. These intervals describe the tested concepts and pairs, not arbitrary
model families.

\subsection{Claim-to-Test Map}

Table~\ref{tab:logic} links each primary estimand to the control result that
would weaken its interpretation. Static and observational measurements are not
used as evidence for the causal claim.

\begin{table*}[t]
\centering
\small
\caption{Claim-to-test map. Controls address the simplest alternative
explanation for each measurement.}
\label{tab:logic}
\begin{tabular}{p{0.20\textwidth}p{0.25\textwidth}p{0.43\textwidth}}
\toprule
Question & Primary estimand & Interpretation would weaken if \\
\midrule
Does static geometry track scale? & Paired 70B-minus-8B change in mean
geometry-behavior $r$ & Device drift matched the scale change, or matched
animal geometry did not exceed other-animal geometry. \\
Does readability share the causal scale change? & Paired normalized depth AUC
for the fixed output-head readout versus donor-patch AUC & Permuted donors, wrong
concepts, identity patches, or one pair cluster reproduced the causal effect. \\
Is timing only relative-depth bookkeeping? & Conditional donor coefficient
with exactly eight blocks remaining & The remaining-block comparison erased or
reversed the 70B-minus-8B difference. \\
Does sequence scoring recover the atomic association? & Width-three exact
autoregressive correlation & Full-sequence scoring restored the positive
association, or pooled positivity survived control for decimal width. \\
\bottomrule
\end{tabular}
\end{table*}

\begin{figure*}[t]
  \centering
  \includegraphics[width=0.96\textwidth]{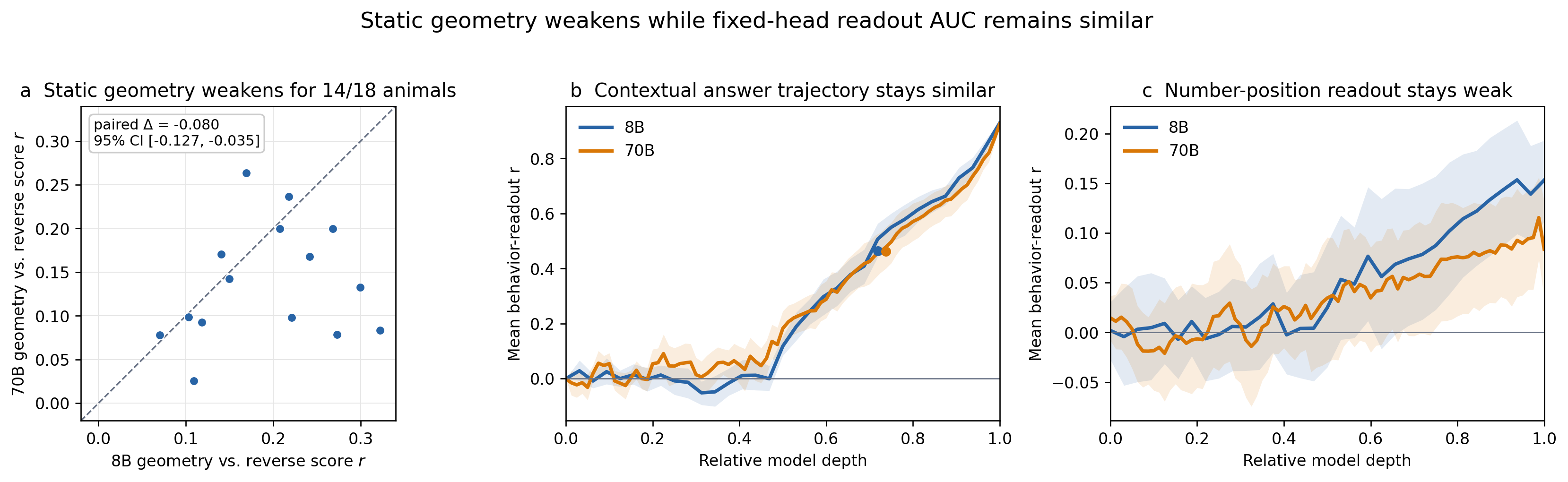}
  \caption{Static geometry and observational readout. (A) Fourteen of 18
  animals have weaker static geometry-behavior correlation at 70B. (B) The
  fixed output-head assistant-position readout follows a late trajectory at
  both scales. (C) Readout at number positions remains weak. Intervals resample
  the same 18 animals.}
  \label{fig:geometry}
\end{figure*}

\begin{figure*}[t]
  \centering
  \includegraphics[width=0.96\textwidth]{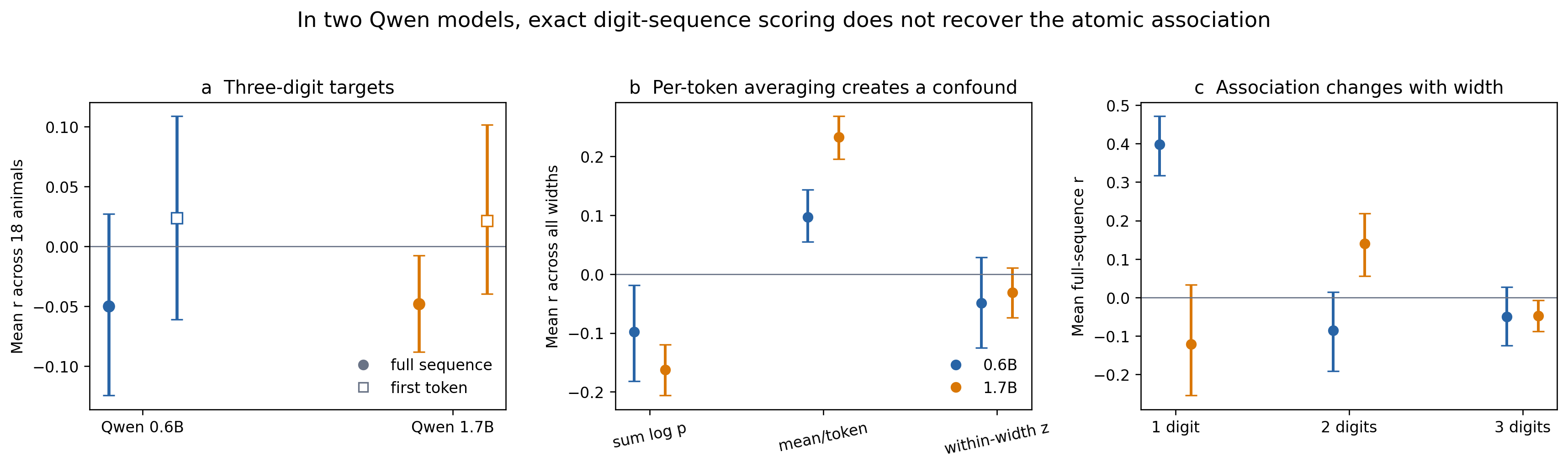}
  \caption{Multi-token scoring in two Qwen models. Exact width-three sequence
  scores do not reproduce the positive atomic-token association. Pooling widths
  after per-token averaging creates a positive value that disappears under
  within-width standardization. Width one contains only ten strings and is
  descriptive.}
  \label{fig:qwen}
\end{figure*}

\begin{table*}[t]
\centering
\caption{Matched Llama-3.1 comparison. Rows report animal means. Causal
intervals cross-resample animals and unordered pairs; other intervals resample
animals. ``Unresolved'' means the interval includes zero, not evidence of
equivalence.}
\label{tab:summary}
\begin{tabular}{lccc}
\toprule
Measurement & 8B & 70B & Paired 70B minus 8B \\
\midrule
Bidirectional behavior, mean $r$ & 0.1087 & 0.0846 & $-0.0241$ [$-0.0563$, $+0.0087$] \\
Static geometry vs. behavior & 0.1877 & 0.1076 & $-0.0802$ [$-0.1271$, $-0.0347$] \\
Geometry specificity & 0.1550 & 0.0458 & $-0.1092$ [$-0.1640$, $-0.0599$] \\
Observational readout AUC & 0.2591 & 0.2688 & $+0.0096$ [$-0.0280$, $+0.0453$] \\
Causal donor AUC & 0.2539 & 0.5397 & $+0.2858$ [$+0.2716$, $+0.2999$] \\
\bottomrule
\end{tabular}
\end{table*}

\begin{table*}[t]
\centering
\small
\caption{Conditional coefficients at every measured block input. The depth-zero
point is analytic. Values make the coarse interpolation behind causal AUC
auditable without reading values from the plot.}
\label{tab:depth}
\begin{tabular}{lrrrrrr}
\toprule
Model and coefficient & 0 & 0.25 & 0.50 & 0.75 & approx. 0.90 & approx. 0.97 \\
\midrule
8B donor $\beta_{\mathrm{donor}}$ & 0 & 0.0012 & 0.0384 & 0.5923 & 0.7799 & 0.9738 \\
8B recipient $\gamma_{\mathrm{recipient}}$ & 1 & 0.9981 & 0.9434 & 0.4350 & 0.2435 & 0.0311 \\
70B donor $\beta_{\mathrm{donor}}$ & 0 & 0.0073 & 0.7728 & 0.9380 & 0.9482 & 0.9844 \\
70B recipient $\gamma_{\mathrm{recipient}}$ & 1 & 0.9888 & 0.1888 & 0.0716 & 0.0610 & 0.0197 \\
\bottomrule
\end{tabular}
\end{table*}

\begin{table}[t]
\centering
\small
\caption{Causal specificity and sensitivity. Wrong-concept entries are ranges
over all 17 circular shifts.}
\label{tab:controls}
\begin{tabular}{@{}lcc@{}}
\toprule
Check & 8B & 70B \\
\midrule
Matched causal AUC & 0.2539 & 0.5397 \\
Wrong-concept AUC & $[-.0406,.0108]$ & $[-.0719,.0145]$ \\
Raw-logit AUC & 0.2668 & 0.5411 \\
Identity-patch max error & 0 & 0 \\
\bottomrule
\end{tabular}
\end{table}

\section{Results}

\subsection{Fixed Geometry Predicts Less; the Readout Change Is Unresolved}

The animal-number behavioral association remains measurable at 70B, but its scale change is
unresolved (Table~\ref{tab:summary}). Mean animal-number correlation changes
from $0.1087$ to $0.0846$; the paired interval includes zero and 6 of 18 animals
increase. Medians are $0.117$ and $0.088$.

Fixed output-vector similarity does change clearly. Its mean correlation with reverse behavior
falls from $0.188$ to $0.108$, a paired change of $-0.080$; 14 of 18 animals
decrease. Matched-minus-other specificity falls by $-0.109$. Repeating 8B on
MPS and CUDA changes the primary statistic by $-0.00083$, much smaller than the
scale contrast.

The observational assistant-position readout does not resolve a corresponding
change (Figure~\ref{fig:geometry}). Normalized AUC is $0.259$
(\CI{$0.236$}{$0.282$}) at 8B and $0.269$
(\CI{$0.235$}{$0.300$}) at 70B. The paired interval for $+0.010$ includes zero.
Mean curves reach half their final values at relative depths $0.719$ and
$0.738$. System-number-position AUC remains low ($0.051$ and $0.036$). The 8B
backend check changes assistant AUC by less than $0.00002$.

A post hoc saved-array sensitivity restricts geometry and readout to the exact
256 causal numbers. Geometry still weakens by $-0.0646$
(\CI{$-0.1243$}{$-0.0033$}), while the readout change remains unresolved at
$+0.0335$ (\CI{$-0.0052$}{$+0.0715$}).

\subsection{The Donor State Controls the 70B Answer Earlier in Relative Depth}

At the five measured depths, mean $\beta_{\mathrm{donor}}$ values at 8B are
$0.001$, $0.038$, $0.592$, $0.780$, and $0.974$. At 70B they are
$0.007$, $0.773$, $0.938$, $0.948$, and $0.984$. At half depth, retained recipient
control is $0.943$ at 8B and $0.189$ at 70B. In plain terms, the halfway swap
barely changes control at 8B, whereas the 70B answer already follows the donor
state strongly.

Normalized causal AUC changes from $0.254$ to $0.540$, with paired difference
$+0.286$; all 18 concepts increase. Descriptively, 14 of 18 show weaker
full-universe static geometry and stronger subset causal AUC. The scale contrast also appears at a
matched remaining-block point: with exactly eight transformer blocks remaining,
donor control is $0.592$ at 8B and $0.948$ at 70B. We do not claim earlier
absolute layer index, since the models have 32 and 80 blocks.

Controls isolate matched donor transmission (Table~\ref{tab:controls}). Every
wrong-concept AUC is far below its matched value. Raw, uncontrasted animal
logits preserve the scale change ($+0.274$, \CI{$+0.259$}{$+0.289$}, 18/18
positive). Leaving out each of
128 unordered pair clusters yields paired changes from $+0.2850$ to $+0.2876$.
Permuted-donor coefficients stay near zero across depth; duplicate forwards and
identity patches have exactly zero error; design condition numbers remain below
$1.25$; and neither direction half drives the result.

\subsection{Whole-Sequence Scoring Does Not Recover the Atomic-Token Association}

For 1,000 width-three Qwen strings, exact full-sequence correlation is $-0.050$
(\CI{$-0.124$}{$+0.027$}) at 0.6B and $-0.048$
(\CI{$-0.088$}{$-0.008$}) at 1.7B. First-token correlations are $+0.024$ and
$+0.021$. Thus scoring every digit in sequence does not recover a positive
atomic-style association in either tested Qwen model. Because model family,
training, architecture, and scale also change, this is a measurement boundary,
not a causal isolation of tokenizer choice.

Pooling widths exposes the normalization confound. Sequence-sum correlations are
$-0.098$ and $-0.162$, whereas per-token averaging reverses them to $+0.097$
and $+0.233$. After standardizing both variables within width, they return to
$-0.049$ and $-0.032$ (Figure~\ref{fig:qwen}). The apparent positive result
comes from differences between the one-, two-, and three-digit groups, not a
positive relationship within those groups. Formally, this is the law of total
covariance:
\begin{align}
\operatorname{Cov}(X,Y)={}&\mathbb{E}_w[\operatorname{Cov}(X,Y\mid w)]\nonumber\\
&+\operatorname{Cov}_w\{\mathbb{E}[X\mid w],\mathbb{E}[Y\mid w]\}.
\end{align}
Here $X,Y$ are the directional scores and $w$ is width. Because per-token
averaging changes width-specific means and width equals token count, a positive
between-width term can overwhelm non-positive within-width associations.

\section{Discussion}

The four measurements answer different questions and yield different scale
trends: behavior persists, fixed output-vector similarity predicts less,
fixed-head readout AUC is unresolved, and donor control shifts earlier in
relative depth. In this setting, ``entanglement'' is therefore not a single
quantity that simply rises or falls with model size.

The geometry-causality split constrains a simple fixed-vector explanation. If
static output-vector similarity were the whole scale-monotonic account, it
should not become less predictive while donor control becomes stronger for 14
of 18 concepts. The results instead distinguish the fixed output vocabulary
from the temporary computation that reaches it. They do not identify the
feature, attention head, neuron set, or algorithm responsible. Full-state
patching shows that the inserted donor state can control the output at the
measured late depths; it does not establish necessity or a unique circuit.
Engineered subspace patches can produce valid end-to-end effects through
dormant pathways \citep{makelov2024subspace}; using natural full states avoids
that particular construction but does not turn intervention-specific control
into mechanism identification.

Readability need not imply causal use
\citep{belinkov2022probing,ma2026plan}. In this channel, static geometry
weakens while the causal trace shows a large timing shift in the opposite
direction.

The Qwen analysis distinguishes sequence log probability, mean log probability
per token, and width-controlled association. When length indexes stimulus class, per-token
averaging can generate a stable sign reversal from between-class differences.
Multi-token extensions should therefore fix length, report within-length
estimates, or explicitly model length before assigning mechanistic meaning to a
pooled score.

\paragraph{Limitations.}
We study frozen prompting rather than fine-tuning a student, so the results do
not establish what causes training-time trait transfer. One same-release Llama
pair cannot establish a universal scaling law. Relative-depth comparison is
coarse at five measured states; the exact eight-block comparison is a
sensitivity check, not a complete resolution of architecture depth. Two small
Qwen models cannot separate tokenizer from architecture, training data, or
scale. Reverse behavior scores the canonical first animal token rather than
marginalizing all synonymous strings. The fixed output-head trace is
observational, and full-residual patching is not a feature-level circuit
intervention. Natural donor vectors come from real source runs, but the hybrid
donor-state/recipient-context computation need not remain on-manifold. Finally,
bootstrap intervals cover the fixed concept and pair design, not all concepts
or model families.

An outcome-blinded external check cautions against interpreting donor timing as
a predictor of student transfer. Against released single-seed outcomes for 16
animals, none of the three frozen in-house measures cleared the fixed
multiplicity-corrected prediction gate. Static geometry was suggestive
($\rho=0.562$, BH $q=0.078$), causal donor AUC was not predictive
($\rho=0.111$, $q=0.687$), and the released steering benchmark was more strongly
associated with transfer ($\rho=0.768$; see Supplement;
\citealp{blank2026steering}). A prospective test could train students across
multiple teachers, seeds, and compatibility regimes, then evaluate static
geometry, observational depth, and causal donor timing as separate predictors.

\section{Conclusion}

In a matched Llama-3.1 comparison, the 70B model's temporary answer-position
state gains donor control earlier, even though fixed animal-number
output-vector similarity predicts behavior less well. Simply reading the
hidden state does not reveal this causal change. Across a Qwen
tokenizer/model-family boundary, correct whole-sequence scoring also fails to
recover the atomic-token association, while naive per-token averaging creates
a positive result driven by number width. Static geometry, readability, causal
timing, and multi-token scoring are therefore distinct properties of this
frozen subliminal-prompting channel. Code and saved summaries accompany the
repository.

\paragraph{AI Assistance.}
Generative AI tools supported implementation and manuscript preparation.

\bibliography{references}

\begin{thebibliography}{23}
\providecommand{\natexlab}[1]{#1}

\bibitem[{{Anonymous Authors}(2026)}]{positional2026}
{Anonymous Authors}. 2026.
\newblock Subliminal Transfer of Positional Biases in Language Models.
\newblock \emph{OpenReview preprint, ICML submission}.
\newblock Preliminary work.

\bibitem[{Belinkov(2022)}]{belinkov2022probing}
Belinkov, Y. 2022.
\newblock Probing Classifiers: Promises, Shortcomings, and Advances.
\newblock \emph{Computational Linguistics}, 48(1): 207--219.

\bibitem[{Belrose et~al.(2023)Belrose, Furman, Smith, Halawi, Ostrovsky,
  McKinney, Biderman, and Steinhardt}]{belrose2023tuned}
Belrose, N.; Furman, Z.; Smith, L.; Halawi, D.; Ostrovsky, I.; McKinney, L.;
  Biderman, S.; and Steinhardt, J. 2023.
\newblock Eliciting Latent Predictions from Transformers with the Tuned Lens.
\newblock \emph{arXiv preprint arXiv:2303.08112}.

\bibitem[{Blank et~al.(2026)Blank, Bhatia, Rajamanoharan, Conmy, and
  Nanda}]{blank2026steering}
Blank, C.; Bhatia, A.; Rajamanoharan, S.; Conmy, A.; and Nanda, N. 2026.
\newblock Subliminal Learning Is Steering Vector Distillation.
\newblock \emph{arXiv preprint arXiv:2606.00995}.

\bibitem[{Brockers et~al.(2026)Brockers, Ventzke, Neuhaus, Hidalgo-Ogalde, and
  Priesemann}]{brockers2026noise}
Brockers, V.~C.; Ventzke, R.~D.; Neuhaus, V.; Hidalgo-Ogalde, B.; and
  Priesemann, V. 2026.
\newblock Learning Through Noise: Why Subliminal Learning Works and When It
  Fails.
\newblock \emph{arXiv preprint arXiv:2605.23645}.

\bibitem[{Chauhan and Shah(2026)}]{chauhan2026covert}
Chauhan, K.; and Shah, A. 2026.
\newblock Covert Trait Propagation Is Representation Alignment: Mechanistic
  Evidence from Hidden-Channel Distillation.
\newblock \emph{arXiv preprint arXiv:2607.04432}.

\bibitem[{Cloud et~al.(2025)Cloud, Le, Chua, Betley, Sztyber-Betley, Hilton,
  Marks, and Evans}]{cloud2025subliminal}
Cloud, A.; Le, M.; Chua, J.; Betley, J.; Sztyber-Betley, A.; Hilton, J.; Marks,
  S.; and Evans, O. 2025.
\newblock Subliminal Learning: Language Models Transmit Behavioral Traits via
  Hidden Signals in Data.
\newblock \emph{arXiv preprint arXiv:2507.14805}.

\bibitem[{Geh et~al.(2024)Geh, Zhang, Ahmed, Wang, and Van
  Den~Broeck}]{geh2024tokenization}
Geh, R.; Zhang, H.; Ahmed, K.; Wang, B.; and Van Den~Broeck, G. 2024.
\newblock Where Is the Signal in Tokenization Space?
\newblock In \emph{Proceedings of the 2024 Conference on Empirical Methods in
  Natural Language Processing}, 3966--3979.

\bibitem[{Geva et~al.(2022)Geva, Caciularu, Wang, and
  Goldberg}]{geva2022vocabulary}
Geva, M.; Caciularu, A.; Wang, K.~R.; and Goldberg, Y. 2022.
\newblock Transformer Feed-Forward Layers Build Predictions by Promoting
  Concepts in the Vocabulary Space.
\newblock In \emph{Proceedings of the 2022 Conference on Empirical Methods in
  Natural Language Processing}, 30--45.

\bibitem[{Grattafiori et~al.(2024)Grattafiori, Dubey, Jauhri
  et~al.}]{grattafiori2024llama}
Grattafiori, A.; Dubey, A.; Jauhri, A.; et~al. 2024.
\newblock The Llama 3 Herd of Models.
\newblock \emph{arXiv preprint arXiv:2407.21783}.

\bibitem[{Lesci et~al.(2025)Lesci, Meister, Hofmann, Vlachos, and
  Pimentel}]{lesci2025tokenisation}
Lesci, P.; Meister, C.; Hofmann, T.; Vlachos, A.; and Pimentel, T. 2025.
\newblock Causal Estimation of Tokenisation Bias.
\newblock In \emph{Proceedings of the 63rd Annual Meeting of the Association
  for Computational Linguistics}, 28325--28340.

\bibitem[{Lieberum et~al.(2023)Lieberum, Rahtz, Kramar, Irving, Shah, and
  Mikulik}]{lieberum2023circuits}
Lieberum, T.; Rahtz, M.; Kramar, J.; Irving, G.; Shah, R.; and Mikulik, V.
  2023.
\newblock Does Circuit Analysis Interpretability Scale? Evidence from Multiple
  Choice Capabilities in Chinchilla.
\newblock \emph{arXiv preprint arXiv:2307.09458}.

\bibitem[{Ma and Rui(2026)}]{ma2026plan}
Ma, N.; and Rui, N. 2026.
\newblock Where's the Plan? Locating Latent Planning in Language Models with
  Lightweight Mechanistic Interventions.
\newblock \emph{arXiv preprint arXiv:2605.07984}.
\newblock Accepted at the ICML Workshop on Mechanistic Interpretability.

\bibitem[{Madl(2026)}]{madl2026channel}
Madl, T. 2026.
\newblock Channel Location Constrains the Auditability of Subliminal Learning.
\newblock \emph{arXiv preprint arXiv:2606.22019}.

\bibitem[{Makelov et~al.(2024)Makelov, Lange, Geiger, and
  Nanda}]{makelov2024subspace}
Makelov, A.; Lange, G.; Geiger, A.; and Nanda, N. 2024.
\newblock Is This the Subspace You Are Looking for? An Interpretability
  Illusion for Subspace Activation Patching.
\newblock In \emph{International Conference on Learning Representations}.

\bibitem[{Meng et~al.(2022)Meng, Bau, Andonian, and Belinkov}]{meng2022rome}
Meng, K.; Bau, D.; Andonian, A.; and Belinkov, Y. 2022.
\newblock Locating and Editing Factual Associations in {GPT}.
\newblock In \emph{Advances in Neural Information Processing Systems},
  volume~35.

\bibitem[{Morgulis and Hewitt(2026)}]{morgulis2026steering}
Morgulis, G.; and Hewitt, J. 2026.
\newblock Subliminal Steering: Stronger Encoding of Hidden Signals.
\newblock \emph{arXiv preprint arXiv:2604.25783}.

\bibitem[{Sanz-Guerrero and von~der Wense(2025)}]{sanz2025length}
Sanz-Guerrero, M.; and von~der Wense, K. 2025.
\newblock Mitigating Label Length Bias in Large Language Models.
\newblock In \emph{Proceedings of the 14th International Joint Conference on
  Natural Language Processing and the 4th Conference of the Asia-Pacific
  Chapter of the Association for Computational Linguistics}, 1404--1420.

\bibitem[{Schrodi et~al.(2026)Schrodi, Kempf, Barez, and
  Brox}]{schrodi2026understanding}
Schrodi, S.; Kempf, E.; Barez, F.; and Brox, T. 2026.
\newblock Towards Understanding Subliminal Learning: When and How Hidden Biases
  Transfer.
\newblock In \emph{International Conference on Learning Representations}.

\bibitem[{Wang et~al.(2026)Wang, Xu, Fang, Yao, Deng, Chen, and
  Zhang}]{wang2026data2behavior}
Wang, M.; Xu, Z.; Fang, J.; Yao, Y.; Deng, S.; Chen, H.; and Zhang, N. 2026.
\newblock From Data to Behavior: Predicting Unintended Model Behaviors Before
  Training.
\newblock \emph{arXiv preprint arXiv:2602.04735}.
\newblock Work in progress.

\bibitem[{Yang et~al.(2025)Yang, Li, Yang et~al.}]{yang2025qwen3}
Yang, A.; Li, A.; Yang, B.; et~al. 2025.
\newblock Qwen3 Technical Report.
\newblock \emph{arXiv preprint arXiv:2505.09388}.

\bibitem[{Zhang and Nanda(2023)}]{zhang2023patching}
Zhang, F.; and Nanda, N. 2023.
\newblock Towards Best Practices of Activation Patching in Language Models:
  Metrics and Methods.
\newblock \emph{arXiv preprint arXiv:2309.16042}.

\bibitem[{Zur et~al.(2025)Zur, Ying, Loftus, {\c{S}}ahin, Yu, Quirke,
  Rott~Shaham, Shapira, Orgad, and Bau}]{zur2025entanglement}
Zur, A.; Ying, Z.; Loftus, A.~R.; {\c{S}}ahin, K.; Yu, S.; Quirke, L.;
  Rott~Shaham, T.; Shapira, N.; Orgad, H.; and Bau, D. 2025.
\newblock Token Entanglement in Subliminal Learning.
\newblock In \emph{Mechanistic Interpretability Workshop at NeurIPS 2025}.

\end{thebibliography}

\end{document}